\def\year{2022}\relax
\documentclass[letterpaper]{article} 
\usepackage{aaai22}  
\usepackage{times}  
\usepackage{helvet}  
\usepackage{courier}  
\usepackage[hyphens]{url}  
\usepackage{graphicx} 
\usepackage{natbib}  
\usepackage{caption} 
\DeclareCaptionStyle{ruled}{labelfont=normalfont,labelsep=colon,strut=off} 
\usepackage{epsfig}
\usepackage{amsmath}
\usepackage{amssymb}
\usepackage{subcaption}
\usepackage{float}
\usepackage{bbm}
\usepackage{comment}
\usepackage{booktabs}
\usepackage{color}

\usepackage{afterpage}

\newcommand\blankpage{%
    \null
    \thispagestyle{empty}%
    \addtocounter{page}{-1}%
    \newpage}

\usepackage{newfloat}
\usepackage{listings}
\floatstyle{ruled}
\newfloat{listing}{tb}{lst}{}
\floatname{listing}{Listing}

\usepackage[ruled,vlined,linesnumbered,noresetcount]{algorithm2e}
\usepackage{algpseudocode}

\DeclareMathOperator*{\argmax}{arg\,max}

\newcommand{\ours}{DULO}
\newcommand{\smallModel}{proxy model}
\newcommand{\utilityModel}{utility model}

\newcommand{\LabeledSet}{\mathcal{L}}
\newcommand{\Unlabeled}{\mathcal{U}}
\newcommand{\utilityFunc}{u}
\newcommand{\A}{\mathcal{A}}
\newcommand{\B}{\mathcal{B}}
\newcommand{\deepset}{DS}
\newcommand{\targetSelectNum}{M}
\newcommand{\selectionBlockSize}{B}
\newcommand{\selectedSet}{S_{selected}}

\newcommand{\indicator}{\mathbbm{1}}

\newcommand{\V}{\mathcal{V}}
\newcommand{\labelingFunc}{f^*}

\newcommand{\dataUtilityFuncBig}{\dataUtilityFunc_{\A, \utilityFunc, \labelingFunc}}

\newcommand{\trainingset}{\LabeledSet_{tr}}
\newcommand{\testingset}{\LabeledSet_{val}}
\newcommand{\featureExtract}{g}

\newcommand{\dataUtilityFunc}{U}
\newcommand{\classifier}{f}

\title{One-Round Active Learning}
\author {

        Tianhao Wang\textsuperscript{\rm 1}\thanks{Partial work done while at Harvard University},
        Si Chen\textsuperscript{\rm 2},
        Ruoxi Jia\textsuperscript{\rm 2} \\
}
\affiliations {
    \textsuperscript{\rm 1} Princeton University \\
    \textsuperscript{\rm 2} Virginia Tech \\
    tianhaowang@princeton.edu, 
    \{chensi, ruoxijia\}@vt.edu
}

\usepackage{bibentry}

\usepackage[switch]{lineno}
\begin{document}

\maketitle

\begin{abstract}
In this work, we initiate the study of \emph{one-round active learning}, which aims to select a subset of unlabeled data points that achieve the highest model performance after being labeled with only the information from initially labeled data points. The challenge of directly applying existing data selection criteria to the one-round setting is that they are not indicative of model performance when available labeled data is limited. We address the challenge by explicitly modeling the dependence of model performance on the dataset. Specifically, we propose \ours, a data-driven framework for one-round active learning, wherein we learn a model to predict the model performance for a given dataset and then leverage this model to guide the selection of unlabeled data. Our results demonstrate that \ours~leads to the state-of-the-art performance on various active learning benchmarks in the one-round setting. 
\end{abstract}

\section{Introduction}
The success of deep learning largely depends on the access to large labeled datasets; yet, annotating a large dataset is often costly and time-consuming.
Active learning (AL) has been the main solution to reducing labeling costs. The goal of AL is to select the most informative points from unlabeled data set to be labeled, so that the accuracy of the classifier increases quickly as the set of labeled examples grows. AL~\cite{fine2002query, freund1997selective, graepel2000kernel, seung1992query, campbell2000query, schohn2000less, tong2001support} typically works in multiple rounds, where each round builds a classifier based on the current training set, and then selects the next example to be labeled. 
This procedure is repeated until we reach a good model or exceeds the labeling budget.

The multi-round nature of the existing AL framework poses great challenges to its wide adaption. Firstly, there lacks specialized platforms where annotators can interact online with the data owner and provide timely feedback. For instance, the existing broadly used data labeling approaches, such as crowdsourcing (e.g., Amazon Mechanical Turk) and outsourcing to companies, do not support a timely interaction between the data owner and annotators. Second, even if one can find such interactive annotators, it is often inefficient to acquire labels one-by-one, especially when each label takes substantial time or when the training process is time-consuming. Various batch-mode AL algorithms~\cite{hoi2006batch,sener2017active,ash2019deep,kirsch2019batchbald, wei2015submodularity, killamsetty2020glister} have been proposed to improve labeling efficiency by selecting multiple examples to be labeled in each round. However, for these algorithms to be effective, they still require many rounds of interaction. Since the examples selected in the current round depend on the examples selected in all previous rounds, existing batch AL algorithms cannot make full advantage of parallel labeling. 

In light of the aforementioned challenges of multi-round AL, we initiate the study of \emph{one-round AL}, where only one round of data selection and label query is allowed. One-round AL starts with very few labeled data points (e.g. a small amount of randomly selected data that are labeled internally or externally). By exploiting these initially labeled data points, a one-round AL strategy selects a large subset of unlabeled data points that could potentially achieve high data utility after being labeled, and then queries their labels \emph{all at once}. Importantly, in the one-round setting, the selected data points can be labeled completely in parallel.

A straightforward approach to one-round AL is to directly use the data selection criteria from the existing multi-round AL methods to perform one-round selection. However, these criteria cannot reliably select unlabeled points resulting in high model performance in one-round setting where only limited labeled data is available. Some of the existing criteria are calculated based on individual predictions of the model trained on the available labeled data~\cite{wei2015submodularity,ash2019badge,wei2015submodularity, killamsetty2020glister}, which tend to be erroneous with very limited labeled data. Others rely on heuristics (e.g., uncertainty~\cite{lewis1994sequential} and diversity~\cite{seung1992query,sener2017active}) that are only weakly correlated with the model performance; worse yet, these heuristics are also calculated from the model trained on the available labeled data, thus becoming even less effective when labeled data points are very limited. 


In this paper, we propose a general one-round AL framework via \textbf{D}ata \textbf{U}tility function \textbf{L}earning and \textbf{O}ptimization (DULO). Our framework is grounded on the notion of \emph{data utility functions}, which map any given set of unlabeled data points to some performance measure of the model trained on the set after being labeled. We propose a natural formulation of the one-round AL problem, which is to seek the set of unlabeled points maximizing the data utility function. 
We adopt a sampled-based approach to \emph{learn} a parametric model that approximates the data utility function, and then leverage greedy algorithms to optimize the learned model. 
We further propose several strategies to improve scalability.
Specifically, to improve efficiency for learning the data utility function for a large model, we propose to learn it based on a simple proxy model instead of the original model; to improve efficiency for optimizing the data utility function for large unlabeled data, we propose to perform greedy optimization over a block of unlabeled data instead of the entire set.

Our contributions can be summarized as follows. \textbf{(1)} We present the first focused study of one-round AL, an important yet underexplored setting, and develop an algorithm based on data utility function learning and optimization. \textbf{(2)} We exhibit several novel empirical observations about data utility functions, including approximate submodularity and their transferrability between different learning algorithms, which provide actionable insights into boosting the efficiency of data utility learning and optimization. \textbf{(3)} We perform extensive experiments and show that \ours~compares favorably to performing state-of-art batch AL algorithms for one round---a natural baseline for one-round AL---on different datasets and models.




\section{Related Work}
Earlier studies of AL focused on the setting where only one example is selected in each round~\cite{fine2002query, freund1997selective, graepel2000kernel, seung1992query, campbell2000query, schohn2000less, tong2001support}. These methods could be time-consuming as they require many rounds of interaction with the data labeler. 
Batch AL~\cite{hoi2006batch} was later proposed to improve the efficiency of label acquisition by querying multiple examples in each round. However, their selection strategy fails to capture the diversity of examples in a batch, thereby often ending up picking redundant examples. 
Recent works on batch AL have attempted to enhance the diversity of a batch. For instance, core-set \cite{sener2017active} uses $k$-center clustering to select informative data points while preserving the geometry of data points. However, it only works well for simple datasets or for models with good architectural prior. 
More recently, BADGE~\cite{ash2019deep} proposes to sample a batch of data points whose gradients have high magnitude and diverse directions.
This method relies on partially-trained models to acquire hypothesized labels and further rank unlabeled data points for selection. 

Existing batch AL methods have utilized submodular acquisition function to enable efficient selection over large unlabeled datasets \cite{wei2014fast,buchbinder2014submodular,bairi2015summarization,wei2014unsupervised,wei2015submodularity,killamsetty2020glister}. However, the submodular acquisition functions in the existing literature usually rely on \emph{hypothesized labels} assigned by the partially-trained classifier. 
For instance, one popular approach, Filtered Active Submodular Selection (FASS) \cite{wei2015submodularity}, selects unlabeled data points by optimizing the data utility functions of simple classifiers such as KNN or Naive Bayes\footnote{The specific utility functions \citet{wei2015submodularity} and \citet{killamsetty2020glister} use are KNN/Naive Bayes' likelihood function on training/validation set.}. The data utility functions for these simple models have analytical form and are proven to be submodular under certain assumptions. 
These submodular utility functions require label information as input, so in order to optimize the submodular acquisition functions over the unlabeled data pool, we need to assign each unlabeled data point with hypothesized label using the partially-trained classifier. 
This can lead to poor selection performance at the initial few rounds as the classifiers' accuracy is usually low at the beginning. Another related state-of-art approach \cite{killamsetty2020glister} performs data selection and model training alternatively using these simple classifiers' submodular utility function and shares the same limitation as FASS and BADGE due to the use of hypothesized labels.
In contrast, our acquisition functions directly estimate the utility of an unlabeled dataset and thus do not require hypothesized labels. 

While recent work on batch AL shows promising results for choosing high-quality data with multi-round selection, our experiments show that with one round, even the state-of-the-art techniques cannot attain satisfactory performance.

\section{Data Utility Functions}
\subsection{Formalization}
A \emph{data utility function} is a mapping from a set of unlabeled data points to a real number indicating the utility of the set. The metrics for utility could be context-dependent. For instance, in AL, the typical goal is to select a set of unlabeled points that can lead to a classifier with the highest test accuracy after labeling. Hence, a natural choice for data utility is the test accuracy of the classifier trained on the selected unlabeled dataset and their corresponding labels.

More formally, we define data utility function in terms of the notion of \emph{learning algorithm} and \emph{metric function}. Let $S_{x}=\{ x_i\}_{i=1}^n$ denote a set of features and $\labelingFunc$ be the ground-truth function that produce the label for any given feature input. A learning algorithm $\A$ is a function that takes a training set $S = \{ (x_i, \labelingFunc(x_i)) \}_{i=1}^n$ and returns a classifier $\classifier$ which approximates $\labelingFunc$. A metric function $\utilityFunc$ takes a classifier as input and outputs its model utility. If we define model utility as test accuracy, the metric function is defined as $\utilityFunc(\classifier, \V) = \frac{1}{|\V|} \sum_{(x, \labelingFunc(x)) \in \V} \indicator[\classifier(x)=\labelingFunc(x)]$ for a test set $\V$ drawn from the same distribution as the training set. However, test set $\V$ is usually not available during the training time. In practice, $\utilityFunc(\classifier, \V)$ is typically approximated by \emph{validation accuracy} $\utilityFunc(\classifier, V)$ where $V$ is a validation set separated from the training set. With a learning algorithm $\A$, a corresponding metric function $\utilityFunc$ and the ground-truth function $\labelingFunc$, we define the data utility function as $\dataUtilityFunc_{\A, \utilityFunc, \labelingFunc}(S_x) = \utilityFunc( \A((S_x, \labelingFunc(S_x))), \V)$, where we slightly abuse the notation and use $\labelingFunc(S_x)$ to denote the set of labels $\{\labelingFunc(x)\}_{x \in S_x}$.

\subsection{Approximate Submodularity}
With the notion of data utility functions, one can formulate an AL problem as an optimization problem that seeks the set of unlabeled data points maximizing the data utility function. However, each evaluation of data utility function requires retraining the model on the input dataset. Thus, we propose to train a parametric model (dubbed a \emph{data utility model} hereinafter) to approximate data utility functions. The AL task will then becomes optimizing the data utility model over the unlabeled data pool. However, the data utility function is a set function; a general set function is not efficiently learnable~\cite{balcan2011learning} and at the same time, optimizing a general set function is NP-hard~\cite{feige1998threshold}. 

\begin{figure}
    \centering
    \includegraphics[width=0.8\columnwidth]{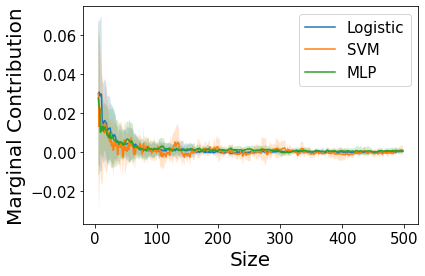}
    \caption{An illustration of ``diminishing return'' property of data utility functions for widely used learning algorithm (trained on USPS dataset, see specific settings in Appendix): 
    given a training set $D$ of size $n$, 
    we randomly sample a data point $j \in D$ at the beginning. 
    We then start with an empty set $S_0 = \emptyset$. For each $i \in \{1, \ldots, n\}$, we randomly sample a data point $k \in D / (S \cup \{j\})$ each time and obtain $S_i = S_{i-1} \cup \{k\}$. Therefore we have $S_0 \subset S_1 \subset \ldots S_{n}$. 
    For each $S_i$, we record the marginal contribution of data point $j$, i.e., $\dataUtilityFunc_{\A, \utilityFunc}(S_i \cup \{j\}) - \dataUtilityFunc_{\A, \utilityFunc}(S_i)$. We repeat the above procedures for 10 different data point $j$ and plot the average of marginal contributions against the size of $S_i$. The procedure exactly matches the definition of submodularity.
    }
    \label{fig:approach-submodularity}
\end{figure}

Fortunately, our empirical studies show that data utility functions exhibit ``approximate'' submodularity. Submodular functions are almost always being characterized by the ``diminishing return'' property. Formally, a set function $f: 2^{V} \rightarrow \mathbb{R}$ returning a real value for any subset $S \subseteq V$ is submodular if $f( \{j\} \cup S)-f(S) \geq f(\{j\} \cup T)-f(T), \forall S \subseteq T, j \in V \setminus T$. 
Figure \ref{fig:approach-submodularity} shows some examples of the marginal contributions of a \emph{fixed} data point versus the sizes of a sequence of data subsets. A clear ``diminishing returns'' phenomenon can be observed from the figure, in which an extra training data point contributes less to model accuracy as the base training set size increases. Moreover, this phenomenon is observed on \emph{almost} every common ML algorithm we test, including complicated models like deep neural networks. Due to space constraint, we defer more examples of such marginal contribution plots on different datasets and models into the Appendix. 

The framework proposed in this paper is greatly inspired by this empirical observation, because it is known that submodular functions can be learned and optimized efficiently. Specifically, it has been theoretically shown that (approximate) submodular functions are PMAC/PAC-learnable with polynomially many independent samples \cite{balcan2011learning,feldman2016optimal,feldman2020tight,wang2021learnability}. Moreover, (approximate) submodular functions can be optimized efficiently by simple greedy algorithms \cite{minoux1978accelerated, horel2016maximization, hassidim2018optimization}.

As a final note, although different learning algorithms' data utility functions consistently exhibit approximate submodularity, the rigorous proof is surprisingly hard. Exact submodularity has thus far been proved for only three simple classes of classifiers--Naive Bayes, Nearest Neighbors, and linear regression, under some strict assumptions \cite{wei2015submodularity,killamsetty2020glister}. We leave the rigorous proof of the approximate submodularity for more widely used models such as logistic regression and deep nets as important future works.

\subsection{Learning Data Utility Model with DeepSets}

Suppose that we have $N$ labeled samples $\LabeledSet = \{(x_i, \labelingFunc(x_i))\}_{i=1}^N$. To learn the data utility model, we first split $\LabeledSet$ into training set $\trainingset$ and validation set $\testingset$, where $\trainingset$ is used for model training and $\testingset$ is used for evaluating the data utility.
Specifically, we train a classifier $\classifier$ on a subset $S \subseteq \trainingset$ and calculate the validation accuracy of $\classifier$ on $\testingset$, which gives us $\utilityFunc(\classifier, \testingset)$. 
If we train the classifier $\classifier$ for multiple times with different subsets $S$, the set $\{(S_x, \utilityFunc(\classifier, \testingset))\}$ could serve as a training set for learning $\dataUtilityFuncBig$, where $S_x$ denote the set of features of $S$. We adopt a canonical model architecture for set function learning--DeepSets~\cite{zaheer2017deep}--as our model for $\dataUtilityFuncBig$. DeepSets is a permutation-invariant type of neural networks, making it suitable for modeling set-valued functions. We then train a DeepSets model $f_{\deepset}$ on $\{ (S_x, \utilityFunc(\classifier, \testingset)) \}$ to approximate $\dataUtilityFuncBig$. 
Algorithm \ref{alg:utilitylearning} summarizes our algorithm for training data utility model. 
For consistency, in the following text, we refer to $\trainingset$ as \emph{labeled training dataset}, and $\testingset$ as \emph{labeled validation dataset}.

\begin{algorithm}[t!]
\SetAlgoLined
\SetKwInOut{Input}{input}
\SetKwInOut{Output}{output}
\Input{$\LabeledSet$ - initially labeled samples, $\A$ - training algorithm for classifier $\classifier$, $\utilityFunc$ - metric function, $\A_{\deepset}$ - training algorithm for  DeepSets model $f_{\deepset}$, $r$ - train-validation split ratio, $T$ - number of classifiers to be trained}
\Output{$f_{\deepset}$ - DeepSets utility model which approximates $\dataUtilityFuncBig$. }

Split $\LabeledSet$ into training set $\trainingset$ and validation set $\testingset$ according to ratio $r$. 

Initialize training set for DeepSets \utilityModel~$S_{\deepset} = \emptyset$.

\For{$t = 1, \dots, T$}{
    Randomly choose a subset $S = (S_x, \labelingFunc(S_x)) \subseteq \trainingset$.
    
    Train classifier $f \leftarrow \A(S)$.
    
    $S_{\deepset} = S_{\deepset} \cup \{ (S_x, u(f, \testingset )) \}$.
}

Train $f_{\deepset} \leftarrow \A_{\deepset}(S_{\deepset})$.

\Return{$f_{\deepset}$}
 \caption{DeepSets Data Utility Learning.}
 \label{alg:utilitylearning}
\end{algorithm}

\section{One-Round AL via \ours}

\subsection{Core Algorithm}
The full algorithm for our proposed one-round AL approach, which we call \ours, is summarized in Algorithm \ref{alg:activelearning}. In a high level, the algorithm first learns the data utility function using the initial labeled dataset and then optimizes the trained data utility model to select the most useful unlabeled data points. 
The learned data utility model $f_{\deepset}$ predicts the test accuracy of the model trained on a given set of unlabeled data \emph{once they are correctly labeled}. The unlabeled data selection problem can be formalized as follows:
\begin{equation}
\argmax_{|S_x|=\targetSelectNum, S_x \subseteq \Unlabeled} f_{\deepset} ( S_x )
\label{eq:maximizingdeepset}
\end{equation} 
where $\targetSelectNum$ is the labeling budget for the selected subset. Since most of the data utility functions are empirically shown to be approximately submodular, we could solve (\ref{eq:maximizingdeepset}) efficiently and approximately using greedy algorithms. 
We choose to apply \emph{stochastic greedy optimization} from \citet{mirzasoleiman2015lazier} to solve (\ref{eq:maximizingdeepset}) because it runs in linear time while provides $1-1/e-\epsilon$ approximation guarantee. 


\subsection{Improving Scalability}
The naive implementation of \ours~is computationally impractical for large models. This is because constructing training samples for the data utility model requires retraining models for many times, which is very slow for large models.
Besides, the clock-time of each DeepSets evaluation increases significantly when the input set sizes grows; hence, optimizing the data utility model over a large unlabeled data pool is expensive. Next, we present strategies to enable \ours~to be scalable to large models and large unlabeled data size.

\paragraph{Scale to Large Models.}
Constructing a size-$T$ training set for learning the data utility model requires retraining the classifier for $T$ times. In experiments, we found that for the data utility function to be learned accurately, we often need to train a classifier thousands of times. The computational costs incurred by this process are acceptable for logistic regression or small CNN models, because the initial labeled dataset is usually very small and it takes minimal time to train one model on it. In our experiment, training 1000 logistic regression models on 300 MNIST images in PyTorch \cite{paszke2019pytorch} only requires around 30 minutes with NVIDIA Tesla K80 GPU, and training 1000 small CNN models on 500 CIFAR-10 images just takes about 4 hours. Besides, since the training of these models is completely independent of each other, constructing training samples for data utility model could be accelerated through parallelization without any communication overhead.

\begin{figure}[H]
    \centering
    \includegraphics[width=0.7\columnwidth]{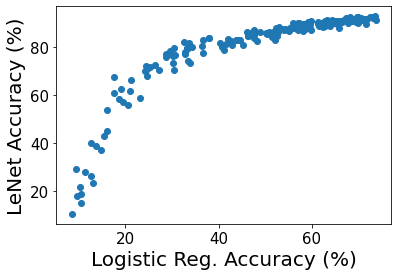}
    \caption{
    Example of the correlation between data utilities for different learning algorithms: for each point in the figure, we randomly sample 2000 MNIST data points, add Gaussian noise to a random portion of the images, and record the accuracies of logistic regression and LeNet trained on the selected samples. Spearman rank correlation here is 0.968. 
    }
    \label{fig:utilitycorrelation}
\end{figure}

However, for relatively large models such as GPT-3 \cite{brown2020language} or very deep ResNets, training on a small dataset multiple times could still be extremely time- or memory-consuming. 
We resolve this problem by utilizing the transferrability of data utility from one learning algorithm to another---data points useful for one learning algorithm are usually useful for other learning algorithms. Such transferrability 
also leveraged in designing efficient data valuation algorithms \cite{jia2019efficient}. 
Figure~\ref{fig:utilitycorrelation} shows an example of such correlation between data utilities. 
As we can see, the logistic regression's accuracies are positively correlated with the LeNet's accuracies, although the LeNet performs much better than the logistic regression. More such examples are provided in the Appendix. 
Based on this observation, we propose to use a \emph{\smallModel} when training original models thousands of times on the initial labeled dataset exceeds time or memory budget. The \smallModel~should be small enough to be trained efficiently, while still achieving observable performance improvement after training. 
When there are too many classes and classification accuracy improvements are hardly noticeable, we can use less stringent utility metrics functions such as Top-5 accuracy. 

\paragraph{Scale to Large Unlabeled Data Pool.}
Another design choice of our one-round AL framework is to deal with the situation when the size of unlabeled dataset is much larger than the labeled set, i.e. $|\Unlabeled| \gg |\trainingset|$. In this case, the generalizability of DeepSets models to giant input sets is questionable, because the largest possible data size that the DeepSets model has seen during training is $|\trainingset|$. Besides, evaluation time of DeepSets will significantly increase for larger input set sizes, which makes the linear-time stochastic greedy algorithm become inefficient in practice. We tackle this challenge by running greedy optimization with a subset $\B \subseteq \Unlabeled$ of an appropriate size $B$. $B$ should be chosen such that the DeepSets data utility model has both good generalizability and efficient evaluation.
We can repeat the above process for $\frac{|\Unlabeled|}{B}$ times and each time, we select $\frac{\targetSelectNum \selectionBlockSize}{|\Unlabeled|}$ data points from each of $\B$. 
We call parameter $B$ as \emph{optimization block size} in Algorithm \ref{alg:activelearning}. Although breaking down the selected data into different blocks, we want to highlight that this design is completely different from batch AL, since we do not require the data points selected in the previous blocks to be labeled before we select data points from the current block. Therefore, the WHILE loop in Algorithm \ref{alg:activelearning} is fully parallelizable.
One may concern that choosing data points independently from each block would fail to capture the interactions between the data selected from different blocks. However, as long as the optimization block size is reasonably large, the missed high-level interactions are negligible. We perform an ablation study for optimization block size $B$ in the Appendix.


\begin{algorithm}
\SetAlgoLined
\SetKwInOut{Input}{input}
\SetKwInOut{Output}{output}
\Input{($\LabeledSet$, $\A$, $\A_{\deepset}$, $\utilityFunc$, $r$, $T$) - same inputs as Algorithm \ref{alg:utilitylearning},  $\Unlabeled$ - unlabeled samples, $\targetSelectNum$ - labeling budget, $\selectionBlockSize$ - optimization block size, $\epsilon$ - precision parameter for stochastic greedy algorithm. }
\Output{$\selectedSet$ - selected set.}

$f_{\deepset} \leftarrow \text{DeepSetsDataUtilityLearning}(\LabeledSet, \A, \A_{\deepset}, \utilityFunc, r, T)$

Initialize $\selectedSet = \emptyset$.

\While{ $|\selectedSet| < \targetSelectNum$ }{
Sample $\B \subseteq \Unlabeled$ of size $\selectionBlockSize$.

$S_x \leftarrow \emptyset$

\While{ $|S_x| < \frac{\targetSelectNum \selectionBlockSize}{|\Unlabeled|}$ }{
Sample $Z \subseteq \B \setminus S_x$ of size $\frac{|\Unlabeled|}{\targetSelectNum} \log(\frac{1}{\log \epsilon})$

Find $e = \argmax_{e \in Z} f_{\deepset}( S_x \cup \{e\})$

$S_x \leftarrow S_x \cup \{e\}$
}

$\selectedSet = \selectedSet \cup S_x$.

$\Unlabeled = \Unlabeled \setminus S_x$.
}

\Return{$\selectedSet$ }
 \caption{\ours~for one-round AL with stochastic greedy optimization}
 \label{alg:activelearning}
\end{algorithm}

\section{Evaluation}
\label{sec:eval}

\subsection{Experimental Settings}
\label{sec:eval-settings}

\subsubsection{Evaluation Protocols}
\label{sec:eval-protocol}
We evaluate the performance of different AL strategies on different types of models and a varied amount of selected data points. Notably, we assess the performance of \ours~for models that are more complicated than \smallModel. 
For example, we use logistic regression as the proxy model for MNIST, while examine the utility of the selected data for LeNet. 
We evaluate the robustness of \ours~on different practical scenarios such as imbalanced or noisy unlabeled datasets. We also test the performance on real-world datasets and provide insights into the selected data. Finally, we perform ablation studies for sample complexity, hyperparameters, and potential variant for \ours. Due to space constraint, we present the results for \ours's sample complexity here and defer other results to Appendix. 

\subsubsection{Datasets \& Implementation Details}
\label{sec:eval-settings-dataset}

We evaluate the robustness of \ours~on class-imbalanced and noisy unlabeled datasets from MNIST \cite{lecun1998mnist} and CIFAR-10 \cite{krizhevsky2009learning}. We also perform experiments on USPS \cite{alpaydin1998optical} and PubFig83 \cite{pinto2011scaling} to demonstrate effectiveness of \ours~on real-world datasets. We summarize the sizes of labeled training set ($\trainingset$), unlabeled set ($\Unlabeled$), labeled validation set ($\testingset$), and \smallModel~in Table \ref{tb:dataset-settings}. All of the $\trainingset$ and $\Unlabeled$ are sampled from the corresponding datasets' original training data. The sampling distribution is uniform unless otherwise specified in the paper. 
For all datasets, we randomly sample 4000 subsets of $\trainingset$ and use the corresponding \smallModel~to generate the training data for utility learning. 
For each subset sampling, we first uniformly randomly generate a real value $\alpha \in [1, 20]$ for each class, and then draw a distribution sample $p$ from Dirichlet distribution $Dir(\alpha_1, \dots, \alpha_{K})$ where $K$ represents the number of classes. We draw a subset with different class sizes proportional to $p$. This sampling design is to ensure the diversity of class distributions in the sampled subsets. 

We set $\epsilon=10^{-5}$ for stochastic greedy optimization. We set the optimization block size $\selectionBlockSize=2000$ in all cases. We select up to half of the data points from  $\Unlabeled$ in the one-round AL. 
For each experiment setting, we repeat \ours~and other baseline algorithms 10 times to obtain the average AL performance and error bars. We defer the dataset descriptions and the architectures, hyperparameters, and training details of proxy models and DeepSets utility models to the Appendix. 

\begin{table}[]
\centering
\resizebox{0.9\columnwidth}{!}{
\begin{tabular}{ccccc}
\toprule
\textbf{Dataset}  & 
\textbf{$|\trainingset|$} & 
\textbf{$|\Unlabeled|$} & 
$|\testingset|$ & \textbf{Proxy Model} \\ \midrule
MNIST    & 300          & 2000          & 300                  & Logistic Reg.                             \\ 
CIFAR-10 & 500          & 20000         & 500              & SmallCNN                                       \\ 
USPS     & 300          & 2000          & 300                  & Logistic Reg.                             \\ 
PUBFIG83 & 500          & 10000         & 500                & LeNet                                       \\ \bottomrule
\end{tabular}
}
\caption{Dataset Settings.}
\label{tb:dataset-settings}
\end{table}

\subsubsection{Baseline Algorithms}
\label{sec:eval-baseline}
We consider state-of-the-art batch AL strategies as our baselines:
\textbf{(1) FASS} \cite{wei2015submodularity}  first filters out the data samples with low uncertainty about predictions. It then selects a subset by first assigning each unlabeled data a hypothesized label with the partially-trained classifier’s prediction and optimizing a Nearest Neighbor submodular function on the unlabeled dataset with hypothesized labels.\footnote{We do not compare to the Naive Bayes submodular function since NN submodular function mostly outperforms NB submodular for non NB models, as demonstrated in the original paper. }
\textbf{(2) BADGE} \cite{ash2019badge} first generates hypothesized labels and selects a subset based on the diverse gradient embedding obtained with the hypothesized samples.
\textbf{(3) GLISTER} \cite{killamsetty2020glister} also generates hypothesized labels and is formulated as a discrete bi-level optimization problem on the hypothesized samples. 
\textbf{(4) Random} is a setting where we randomly select a subset from the unlabeled datasets.

FASS, BADGE, GLISTER are initially designed for multi-round AL, but they can be trivially extended to the one-round setting by limiting the number of data selection rounds to be 1. Since these methods have been shown to outperform techniques like uncertainty sampling \cite{settles2009active} and coreset-based approaches \cite{sener2017active}, we do not compare them in this work. 
We test these baselines with open-source implementation\footnote{https://github.com/decile-team/distil}. The hyperparameter settings for these baselines are deferred to the Appendix.

\subsection{Experiment Results}
\label{sec:eval-results}


We evaluate the performance of \ours~for one-round AL in different dataset settings, including class-imbalanced, noisy, and real-life datasets. For MNIST, we show the accuracies of both logistic regression (MNIST's \smallModel) and LeNet model on the selected data. For CIFAR-10, since the accuracy of its \smallModel~is relatively low (around 30\%) in general, we only show the data selection results for the target CNN model. For USPS dataset, we test the AL performance on logistic regression (USPS's \smallModel) and SVM. For PubFig83, since it has 83 classes, we show the top-5 accuracies of LeNet model (PubFig83's \smallModel) and a CNN model. 

\newcommand{\wid}{0.495}

\begin{figure}[t]
    \centering
    \includegraphics[width=\columnwidth]{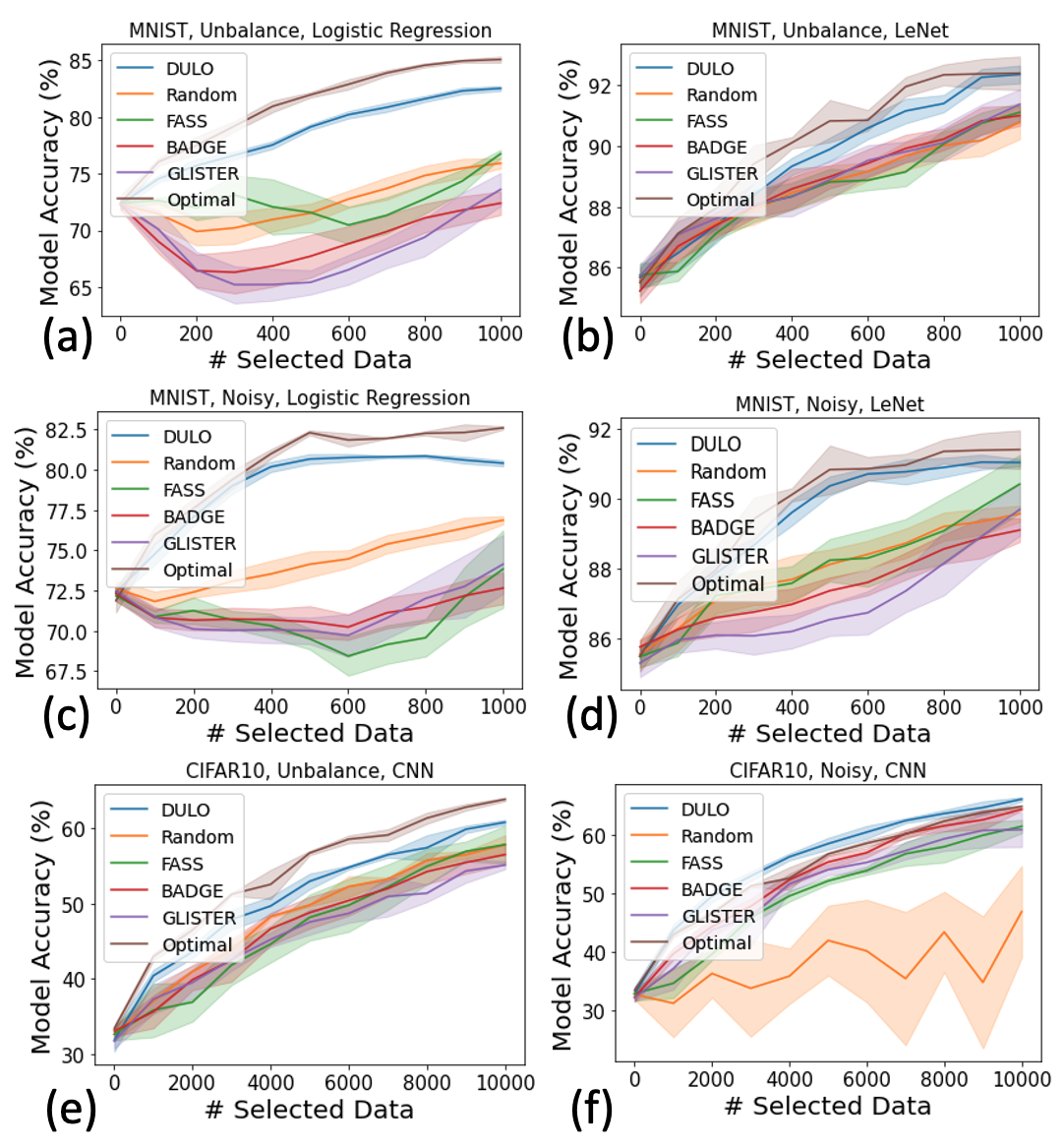}
    \caption{Target models' accuracies vs. the number of selected samples 
    for different settings. Standard errors are shown by shaded regions. The optimal curves are the mean accuracies of balanced/noiseless datasets. }
    \label{fig:activelearning-mnistandcifar}
\end{figure}

\paragraph{Imbalanced Dataset.}
We artificially generate class-imbalance for MNIST unlabeled dataset by sampling 55\% of the instances from one class and the rest of the instances uniformly across the remaining 9 classes. For CIFAR-10's unlabeled dataset, we sample 50\% of the instances from two classes, 25\% instances from another two classes, and 25\% instances from the remaining 6 classes. 
The results for MNIST class-imbalance setting are shown in Figure \ref{fig:activelearning-mnistandcifar} (a) and (b), which are the accuracies of logistic regression (MNIST's \smallModel) and LeNet on the selected datasets, respectively. The results demonstrate that \ours~significantly outperforms the other baselines. By examining the data points selected by different strategies, we found that \ours~can select more balanced dataset, thereby leading to higher performance. We note that \ours's improvement is smaller for LeNet, likely because Convolutional Neural Network is more robust to imbalanced training data than logistic regression. 
Figure \ref{fig:activelearning-mnistandcifar} (e) shows the performance of class-imbalance one-round AL on CIFAR-10. Again, \ours~consistently outperforms other baselines. 

\begin{figure}[t]
    \centering
    \includegraphics[width=0.9\columnwidth]{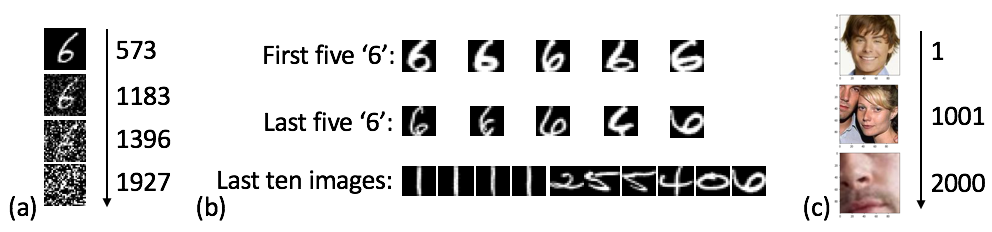}
    \caption{Selection ranks for different images in a greedy optimization block. (a) MNIST (b) USPS (c) PubFig83.}
    \label{fig:mnist-usps-pubfig-rank}
\end{figure}

\paragraph{Noisy Dataset.}
To create noisy datasets, we inject Gaussian noise into images in random subsets of data. The noise scale refers to the standard deviation of the random Gaussian noise.
For MNIST, we add each of the noise scales of 0.25, 0.6, and 1.0 to 25\% unlabeled data (i.e., 75\% of the unlabeled data are noisy and there are 3 different noisy levels). 
For CIFAR-10, we add noise to 25\% unlabeled data points with noise scale 1.0. Figure \ref{fig:activelearning-mnistandcifar} (c) (d) and (f) show the results for noisy data settings. We observe that all three baselines perform very differently on MNIST and CIFAR-10. In contrast, \ours~outperforms other baselines by more than 1\% for every data selection size, and is very close to, or sometimes even better than the optimal curves, which are calculated by taking an average over accuracies of noiseless datasets. 
To get more insights into the selection process, we inject an MNIST image together with its three noisy variants of different noise scales into the MNIST unlabeled dataset and examine their selection order, shown in Figure \ref{fig:mnist-usps-pubfig-rank} (a).
We can see that our strategy tends to select clean, high-quality images early and is able to sift out noisy data.

\begin{figure*}[ht]
    \centering
    \includegraphics[width=0.9\textwidth]{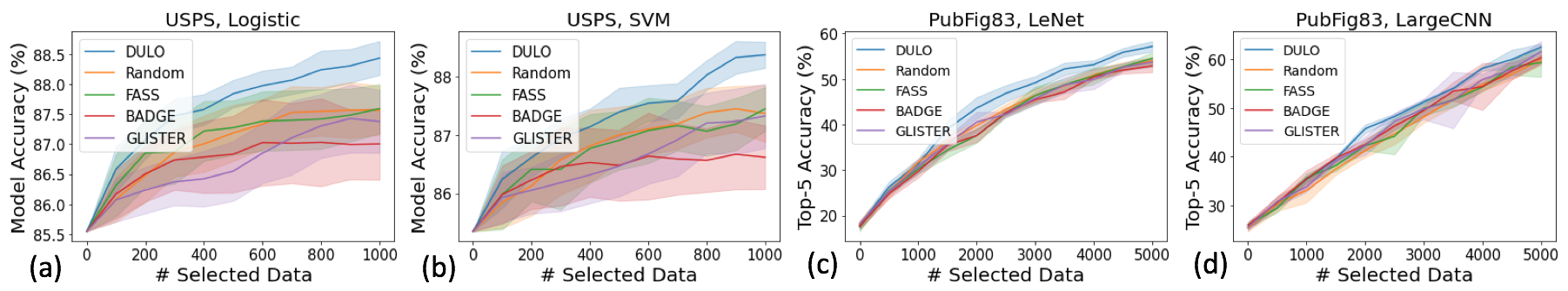}
    \caption{One-round AL on Real-life Datasets.}
    \label{fig:real-life}
\end{figure*}

\begin{figure}[h]
    \centering
    \includegraphics[width=0.9\columnwidth]{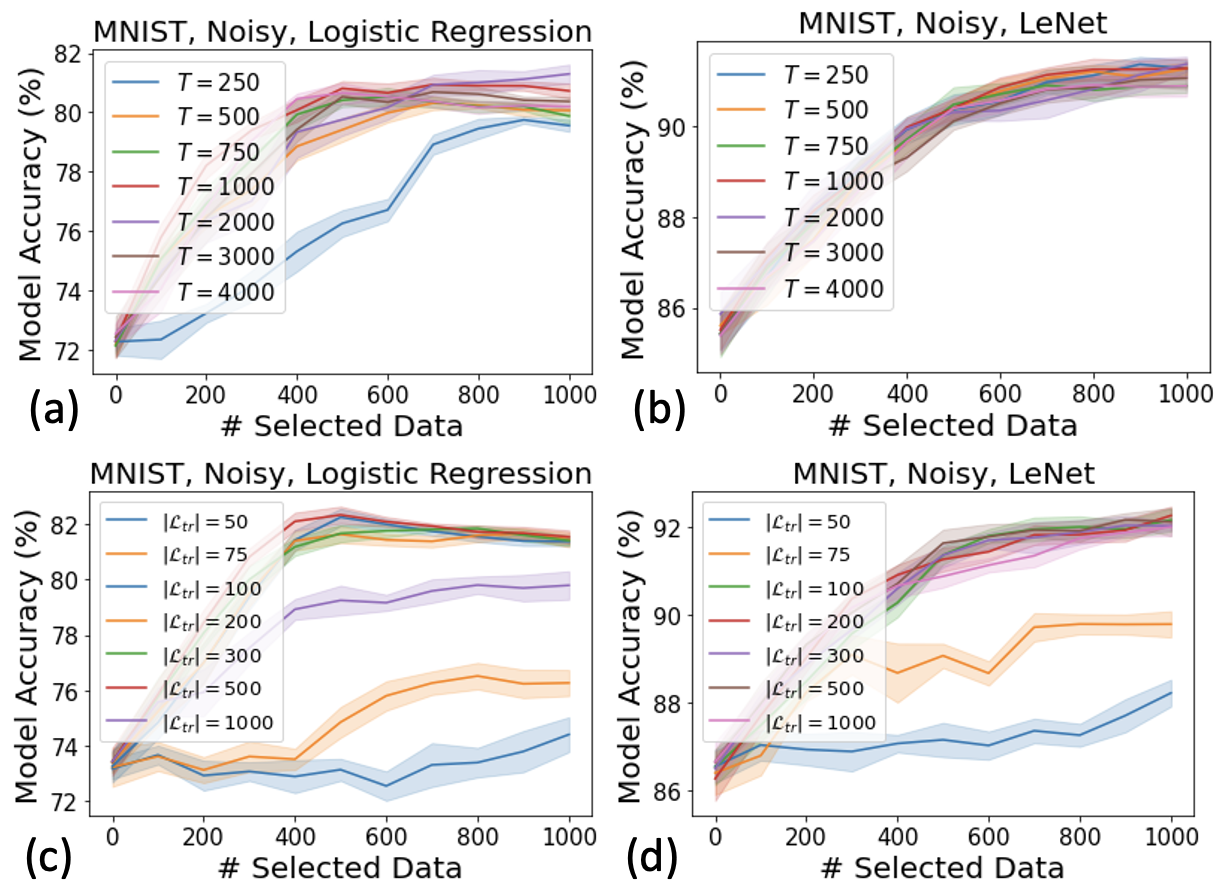}
    \caption{
    (a) and (b) studies the subset complexity for utility learning. (c) and (d) studies data complexity for utility learning. 
    }
    \label{fig:ablation}
\end{figure}

\paragraph{Real-life Datasets.}
We study the effectiveness of different strategies on real-life datasets, including USPS and PubFig83 datasets, which contain some natural variations in image quality.
Figure \ref{fig:real-life} (a) and (b) show the AL performance on USPS dataset with logistic regression (\smallModel) and support vector machine. For both cases, we can see that \ours~consistently outperforms all other baselines. Figure \ref{fig:mnist-usps-pubfig-rank} (b) illustrates the images selected in different ranks by \ours. We can see that the points selected earlier have higher image quality; especially, the last five selected `6' digit images tend to be blurry and some could easily be confused with other digits.
The last ten selected images contain four `1's, which is because USPS is a class-imbalanced dataset, and there are more `1's than other classes. Figure \ref{fig:real-life} (c) and (d) show the performance on PubFig83 dataset with LeNet (\smallModel) and a CNN model. Again, for both cases, we can see that \ours~consistently outperforms all other baselines. Figure \ref{fig:mnist-usps-pubfig-rank} (c) shows the rankings of 3 images within an optimization block. Interestingly, the early face images selected by \ours~contain complete face features, while the later ones either contain more irrelevant features or are corrupted.


\paragraph{Sample Complexity of Data Utility Learning. } 
There are two kinds of sample complexity involved in the DeepSets-based utility model learning: (1) \emph{data complexity} which is the size of labeled training dataset $|\trainingset|$, and (2) \emph{subset complexity} which is the number of subsets $T$ for the initially labeled dataset we sample to train proxy models. 
Both quantities need to be appropriately large to train a utility model that can generalize well to unseen subset and capture the interactions between different data points.
Figure \ref{fig:ablation} (a) and (b) show the one-round AL performance for DeepSets utility models trained with different amount of subset samples. We find that the performance is nearly optimal as long as the number of subset samples is above 500. This is not a large number: training 1000 logistic regression models on MNIST on 300 data points in PyTorch only requires around 30 minutes with NVIDIA Tesla K80 GPU. Since CNN models are more effective on MNIST, even the data utility model trained with only 250 samples is able to achieve comparable performance to the optimal accuracy for learning LeNet. 

Figure \ref{fig:ablation} (c) and (d) show the one-round AL performance for DeepSets utility models trained with different amount of labeled training samples $|\trainingset|$. Note that here the largest possible training samples for the proxy model varies, but we still train the classifiers with 300 labeled training samples together with selected data points.
We fix the block size and number of subsets sampled for DeepSets training in the experiments. 
As we can see, near-optimal performance could be achieved even when there are only 100 labeled training data points. When $|\trainingset|$ is too large, the trained DeepSets model’s performance might degrade due to insufficient training samples compared with input dimensions. When $|\trainingset|$ is too small, the generalizability of DeepSets utility model to large sets would be worse and at the same time, the block size has a small upper limit, which can capture less data interaction.
A practical guideline for setting the size of initial labeled dataset is in the level of hundreds.


\section{Conclusion and Limitations}
In this paper, we identify an important AL setting where only one round of label query is allowed. The simple extension of existing multi-round batch AL algorithms to one-round cannot maintain descent performance. This work proposes a general framework for one-round AL via data utility function learning and optimization. Our evaluation shows that it outperforms existing baselines across various datasets and models. 

There are two assumptions underlying our approach: the labels of the initial labeled dataset is correct and the labeled and unlabeled datasets should share the same set of labels. Extending \ours~to noisy initial labeled data and varied label settings would be interesting future directions. 
Moreover, we would like to complement our currently empirical-only results of approximately submodular data utility and correlated data utility with rigorous analysis, which could provide more guidance on utility model training and proxy model section strategy. 
Scalability is another limitation of our framework. While the evaluation section shows that both the sample and data complexity for utility learning is acceptable, \ours~could still be ineffective for data of very high dimensions because limited initial labeled set does not allow training data utility models with reasonable performance based on proxy models. Further improving the scalability of \ours~through some efficient approximation heuristics of data utility functions would be interesting future works. 


\newpage

\bibliography{ref}

\afterpage{\blankpage}

\newpage

\appendix

\section{Additional Experiment Settings}

\subsection{Figure 1 in the Main Text}
In Figure 1 of the maintext, we use USPS dataset. The MLP we use has 1 hidden layer with 128 neurons, trained with learning rate $10^{-3}$ and batch size 32. 


\subsection{Details of Datasets Used in Section 5}

\paragraph{MNIST.}
MNIST consists of 70,000 handwritten digits. The images are $28 \times 28$ grayscale pixels. 

\paragraph{CIFAR-10.}
CIFAR-10 consists of 60,000 3-channel images in 10 classes (airplane, automobile, bird, cat, deer, dog, frog, horse, ship and truck). Each image is of size $32 \times 32$. 

\paragraph{USPS.}
USPS is a real-life dataset of 9,298 handwritten digits. The images are $16 \times 16$ grayscale pixels. The dataset is class-imbalanced with more $0$ and $1$ than the other digits.

\paragraph{PubFig83.}
PubFig83 is a real-life dataset of 13,837 facial images for 83 individuals. Each image is resized to $32 \times 32$. 

\subsection{Implementation Details}
A small CNN model is used as the proxy model for CIFAR-10 dataset, which has two convolutional layers, each is followed by a max pooling layer and a ReLU activation function. LeNet model is the proxy model for PubFig83 dataset, and we also used it to test the AL performance for MNIST in the experiment. LeNet is adapted from \cite{lecun1998gradient}, which has two convolutional layers, two max pooling layers and one fully-connected layer. A CNN model is used to evaluate AL performance on CIFAR-10 and PubFig83, which has six convolutional layers, and each of them is followed by a batch normalization layer and a ReLU layer. We use Adam optimizer with learning rate $10^{-3}$, mini-batch size 32 to train all of the aforementioned models for 30 epochs, except that we train LeNet for 5 epochs when using it for testing AL performance on MNIST. We also use the support vector machine (SVM) to evaluate AL performance on the USPS dataset. We implement SVM with scikit-learn \cite{pedregosa2011scikit} and set the L2 regularization parameter to be $0.1$. 

A DeepSets model is a function $f_{\deepset}(S) = \rho ( \sum_{x \in S} \phi(x) )$ where both $\rho$ and $\phi$ are neural networks. 
In our experiment, both $\phi$ and $\rho$ networks have 3 fully-connected layers. For MNIST and USPS, we set the number of neurons to be 128 in each hidden layer. For CIFAR-10 and PubFig83, we set the number of neurons in every hidden layer to be 512. 
We set the dimension of set features (i.e., the output of $\phi$ network) for DeepSets models to be 128 in all experiments, except for USPS dataset the set feature number is 64. We use the Adam optimizer with learning rate $10^{-5}$, mini-batch size of 32, $\beta_1=0.9$, and $\beta_2=0.999$ to train all of the DeepSets utility models. 

In terms of the baseline batch AL approaches, we set $\beta$ in FASS to be the size of unlabeled dataset after parameter tuning. We set the learning rate in GLISTER to be 0.05, following the original paper. 


\section{Additional Experiment Results}

\subsection{Diminishing Return}
\begin{figure}[h]
    \centering
    \begin{subfigure}[b]{0.22\textwidth}
        \includegraphics[width=\textwidth]{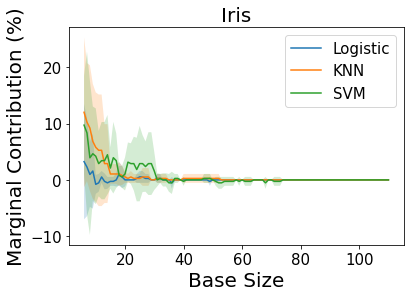}
        \caption{}
        \label{fig:submodularity-iris}
    \end{subfigure}
    \begin{subfigure}[b]{0.22\textwidth}
        \includegraphics[width=\textwidth]{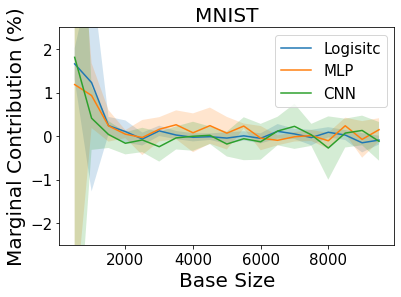}
        \caption{}
        \label{fig:submodularity-mnist}
    \end{subfigure}
    \caption{
    More examples of the ``diminishing return'' property of data utility functions for widely used learning algorithms: (a) is trained on Iris dataset, and (b) is trained on MNIST dataset. The marginal contribution is computed in terms of test accuracy. The sampling procedure is the same as what is described in the Caption of Figure \ref{fig:approach-submodularity}, except for MNIST dataset (Figure (b)), since the size of the training set $D$ is large, at each iteration we randomly sample 500 data points $K \in D / (S \cup \{j\})$ each time and construct $S_i = S_{i-1} \cup \{K\}$. 
    The MLP we use has 1 hidden layer with 128 neurons. The CNN model has two convolutional layers, each is followed by a max pooling layer and a ReLU activation function. Both models are trained with learning rate $10^{-3}$ and batch size 32. 
    }
    \label{fig:submodularity}
\end{figure}

Figure \ref{fig:submodularity} shows more examples of the ``diminishing return'' property of data utility functions for commonly used models and datasets. 
These figures are produced by computing the average marginal contribution of a \emph{fixed} data point to different base sample sets. As we can see, for all cases we consider, the marginal contribution decreases as the size of the base sample sets increase, which matches the definition of submodularity.

\subsection{Utility Transferrability}
\begin{figure*}[t]
    \centering
    \includegraphics[width=2\columnwidth]{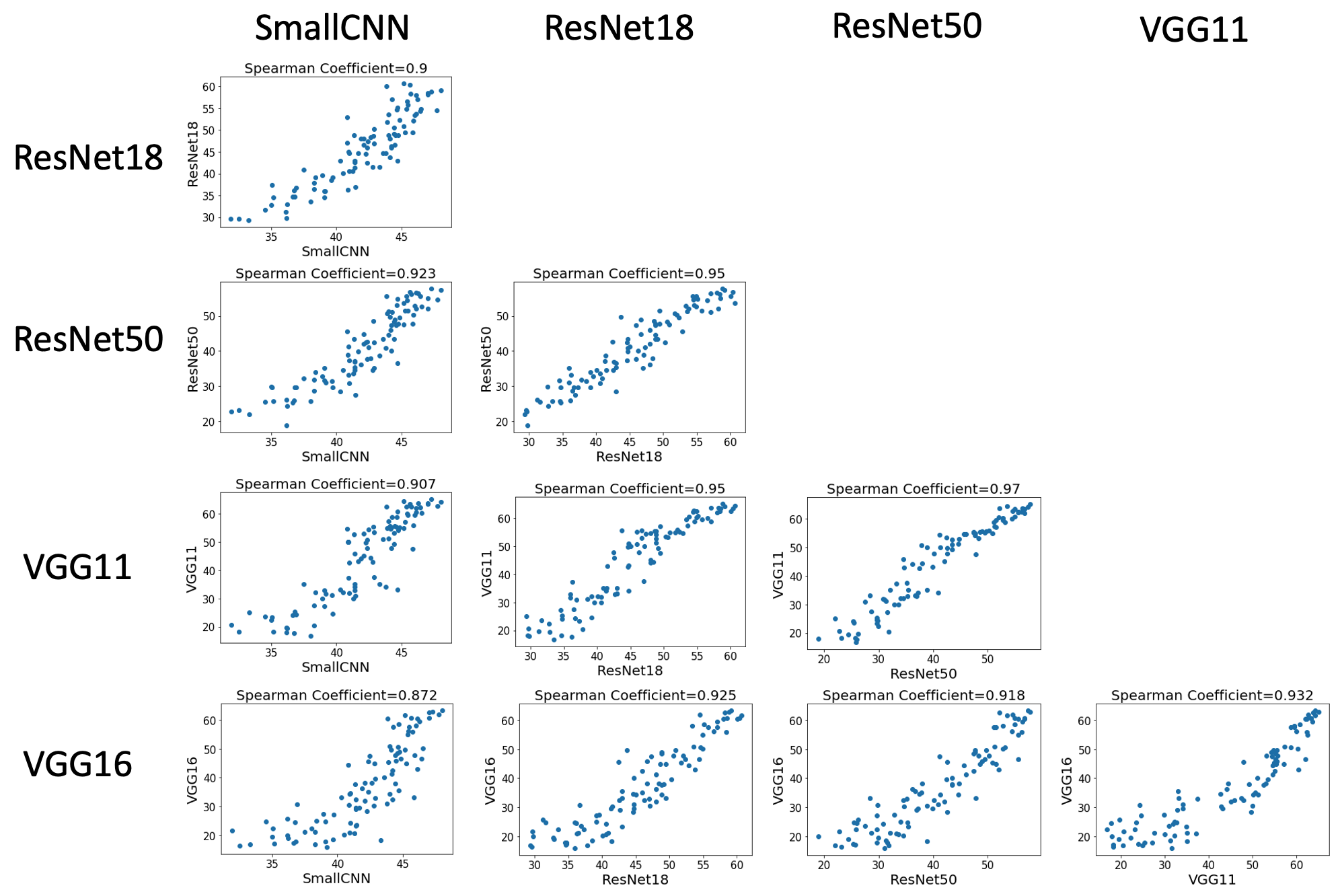}
    \caption{Model Utility Transferrability}
    \label{fig:ablation-transferrability}
\end{figure*}

Figure \ref{fig:ablation-transferrability} shows more examples of the transferrability of data utilities between different learning algorithms. The dataset here we use is CIFAR-10. For each point in the figures, we randomly sample 5000 CIFAR-10 images and add Gaussian noise to a random portion of the images. We then record the test accuracies of different model architectures trained on the dataset. As we can see, for all cases we tested, the Spearman's rank correlation coefficients are high in general, which means that the data utilities are strongly positively correlated. This further illustrates the validity of proxy model approach. However, while the transferrability of data utility across different learning algorithms has important applications, its theoretical analysis seem to be a difficult problem as discussed in \citet{wu2021towards} and it's certainly an important future research direction. 

\subsection{Optimization Block Size}
\begin{figure}[h]
    \centering
    \includegraphics[width=\columnwidth]{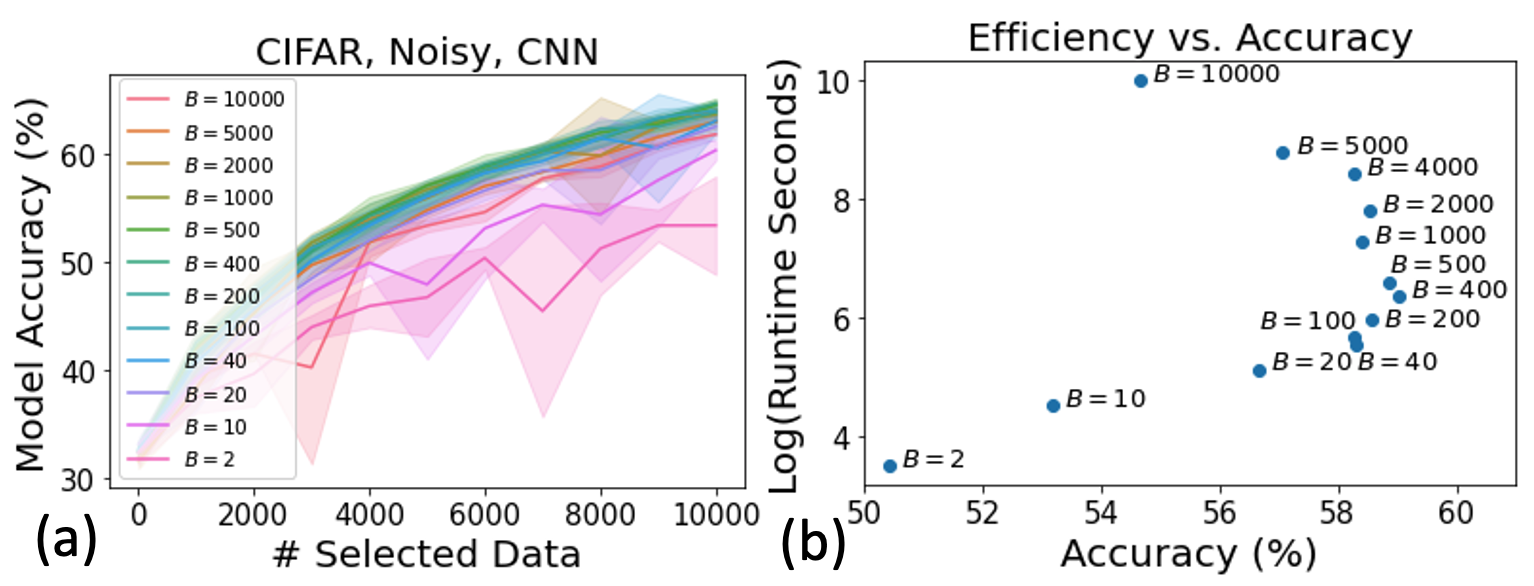}
    \caption{
    Ablation Study: (a) compares \ours's performance with different optimization block sizes, and (b) studies the relationship between efficiency and AL performance for different block sizes when the number of selected data is 6000.
    }
    \label{fig:ablation-blocksize}
\end{figure}

Figure \ref{fig:ablation-blocksize} (a) shows the one-round AL performance with different optimization block sizes $\selectionBlockSize$. 
As we can see, both too small and too large block size can degrade the selected data's utility. When $\selectionBlockSize$ is too small, DeepSets fails to capture interactions between data points selected from different blocks. 
When $\selectionBlockSize$ is too large, DeepSets has large generalization error.


Although the stochastic optimization algorithm runs in $O(|\Unlabeled|)$ regardless of the block size (without parallelization), individual DeepSets evaluation time increases significantly with a larger input set. Figure \ref{fig:ablation-blocksize} (b) shows the runtime vs accuracy plot with different block sizes. We can see that when $\selectionBlockSize$ is too large, it suffers from both poor utility and inefficiency. However, we find a wide range of $\selectionBlockSize$ that achieves both good efficiency and high data utility (the points located on the right of the figure), which is around $|\trainingset|$ ($|\trainingset|=500$ for CIFAR-10 in our setting).
This servers as a heuristic choice of $\selectionBlockSize$. In this range, the block size is small enough so that DeepSets models generalize well on input sets of block sizes, while also being large enough so that the utility model could capture most of the data interactions. 

\begin{figure*}[h]
    \centering
    \begin{subfigure}[b]{0.24\textwidth}
        \includegraphics[width=\textwidth]{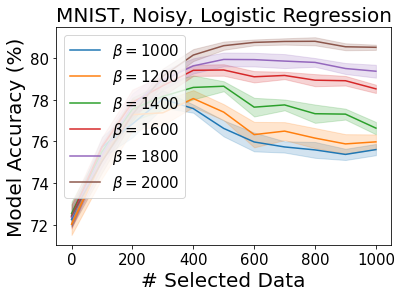}
        \caption{}
        \label{fig:uncertainty-noisy-logistic}
    \end{subfigure}
    \begin{subfigure}[b]{0.24\textwidth}
        \includegraphics[width=\textwidth]{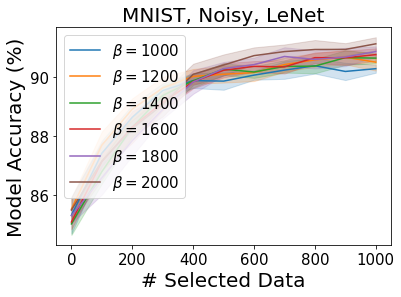}
        \caption{}
        \label{fig:uncertainty-noisy-cnn}
    \end{subfigure}
    \begin{subfigure}[b]{0.24\textwidth}
        \includegraphics[width=\textwidth]{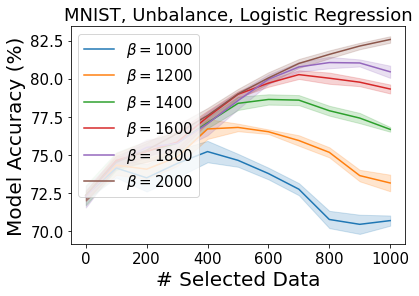}
        \caption{}
        \label{fig:uncertainty-unbalance-logistic}
    \end{subfigure}
    \begin{subfigure}[b]{0.24\textwidth}
        \includegraphics[width=\textwidth]{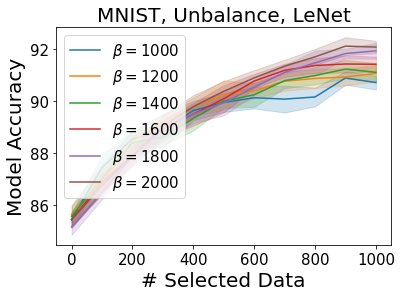}
        \caption{}
        \label{fig:uncertainty-unbalance-cnn}
    \end{subfigure}
    \caption{Ablation Study: the performance of \ours~when we run greedy optimization on the $\beta$ most uncertain samples of the current classifiers. }
    \label{fig:ablation-uncertainty}
\end{figure*}

\subsection{Combining with Uncertainty Sampling}
Traditionally, AL techniques like uncertainty sampling (US) and query by committee (QBC) have shown great promise in several domains of machine learning \cite{settles2009active}. 
However, in the task of multi-round batch AL, naive uncertainty sampling fails to capture the interactions between selected samples. Simply choosing the most uncertain samples may lead to a selected set with very low diversity. 
Filtered Active Submodular Selection (FASS) \cite{wei2015submodularity} combines the uncertainty sampling method with a submodular data subset selection framework. 
Specifically, at every round $t$, FASS first selects a candidate set of $\beta_t$ most uncertain samples among unlabeled data, and then runs greedy optimization on an appropriate submodular objective (with the hypothesized labels assigned by the current model). 
It is natural to ask whether we should also combine \ours~with uncertainty sampling by only performing greedy optimization on the most uncertain samples. 
We evaluate the performance of \ours~when we first filter out data points that the current classifier has high confidence about, and preserve a candidate set of $\beta$ most uncertain samples on which we run stochastic greedy optimization. 
We show the performance of different values of $\beta$ on class-imbalanced and noisy MNIST dataset in Figure \ref{fig:ablation-uncertainty}, where $\beta=|\Unlabeled|=2000$ coincides with the setting of the original \ours~algorithm. As we can see, for all settings studied in our experiments, $\beta=2000$ consistently outperforms other smaller values of $\beta$. Hence, using uncertainty sampling to pre-process unlabeled samples does not seem to lead to better performance in our one-round AL framework.
We conjecture that this is because we only have very few labeled samples at the beginning. In that case, the performance of the initial classifiers is pretty poor and their uncertainty outputs are not informative.

\end{document}